\documentclass[conference]{IEEEtran}
\IEEEoverridecommandlockouts

\usepackage{cite}
\usepackage{amsmath,amssymb,amsfonts}
\usepackage{algorithmic}
\usepackage{textcomp}
\usepackage{mathptmx}
\usepackage{epsfig,utfsym,graphicx,subfigure}
\usepackage{booktabs,multirow,multicol}
\usepackage{xcolor,colortbl}
\def\BibTeX{{\rm B\kern-.05em{\sc i\kern-.025em b}\kern-.08em
    T\kern-.1667em\lower.7ex\hbox{E}\kern-.125emX}}

\begin{document}
\title{Exploring Interactive Semantic Alignment for Efficient HOI Detection with Vision-language Model\\
\thanks{\IEEEauthorrefmark{1}Corresponding author (email: hyang@sjtu.edu.cn).}
}
\author{
\IEEEauthorblockN{Jihao Dong, Renjie Pan, Hua Yang\IEEEauthorrefmark{1}}
    \IEEEauthorblockA{
    Institute of Image Communication and Network Engineering, Shanghai Jiao Tong University, China\\
    Shanghai Key Lab of Digital Media Processing and Transmission, China\\
    China MoE Key Lab of Artificial Intelligence, AI Institute, Shanghai Jiao Tong University, China\\
    }
}

\maketitle

\begin{abstract}
Human-Object Interaction (HOI) detection aims to localize human-object pairs and comprehend their interactions. Recently, two-stage transformer-based methods have demonstrated competitive performance. However, these methods frequently focus on object appearance features and ignore global contextual information. Besides, vision-language model CLIP which effectively aligns visual and text embeddings has shown great potential in zero-shot HOI detection. Based on the former facts, We introduce a novel HOI detector named ISA-HOI, which extensively leverages knowledge from CLIP, aligning interactive semantics between visual and textual features. We first extract global context of image and local features of object to Improve interaction Features in images (IF). On the other hand, we propose a Verb Semantic Improvement (VSI) module to enhance textual features of verb labels via cross-modal fusion. Ultimately, our method achieves competitive results on the HICO-DET and V-COCO benchmarks with much fewer training epochs, and outperforms the state-of-the-art under zero-shot settings.
\end{abstract}


\section{Introduction}
Human-object interaction (HOI) detection is a challenging image understanding problem in computer vision which ultimately obtaining a series of HOI triplets in the format of $<$human, verb, object$>$. Based on this definition, the traditional HOI detection methods can be summarized as two-stage methods~\cite{scg,stip,upt,pvic} and one-stage methods~\cite{qpic,cdn,gen,hoiclip}. One-stage methods detect HOI triplets simultaneously, resulting in high computational costs and slow convergence. In contrast, two-stage methods first utilize pre-trained object detectors to detect humans/objects and subsequently recognize interactions, which solely concentrate on interaction recognition and have demonstrated superior performance by fully leveraging the capabilities of each module.

In recent years, with the research on transformer mechanisms~\cite{attention}, query-based transformer has gained widespread adoption in two-stage methods. STIP~\cite{stip} designed a novel structure-aware transformer to exploit inter-interaction and intra-interaction prior knowledge. UPT~\cite{upt} exploited both unary and pairwise representations of the human and object instances which have complementary properties. PViC~\cite{pvic} introduced box pair positional embeddings as spatial guidance to mine relevant contextual information. The methods mentioned above commonly use appearance features from detectors and manually crafted spatial features to construct the feature representation of human-object pairs, thus \textbf{overlooking the global contextual information} to differentiate interactions. Additionally, in cases where actions exhibit significant spatial patterns, such as \textit{ride}, spatial features dominate the interaction understanding. This might result in the model making excessively confident erroneous predictions.

\begin{figure}[t] 
      \centering
	 \includegraphics[width=3in]{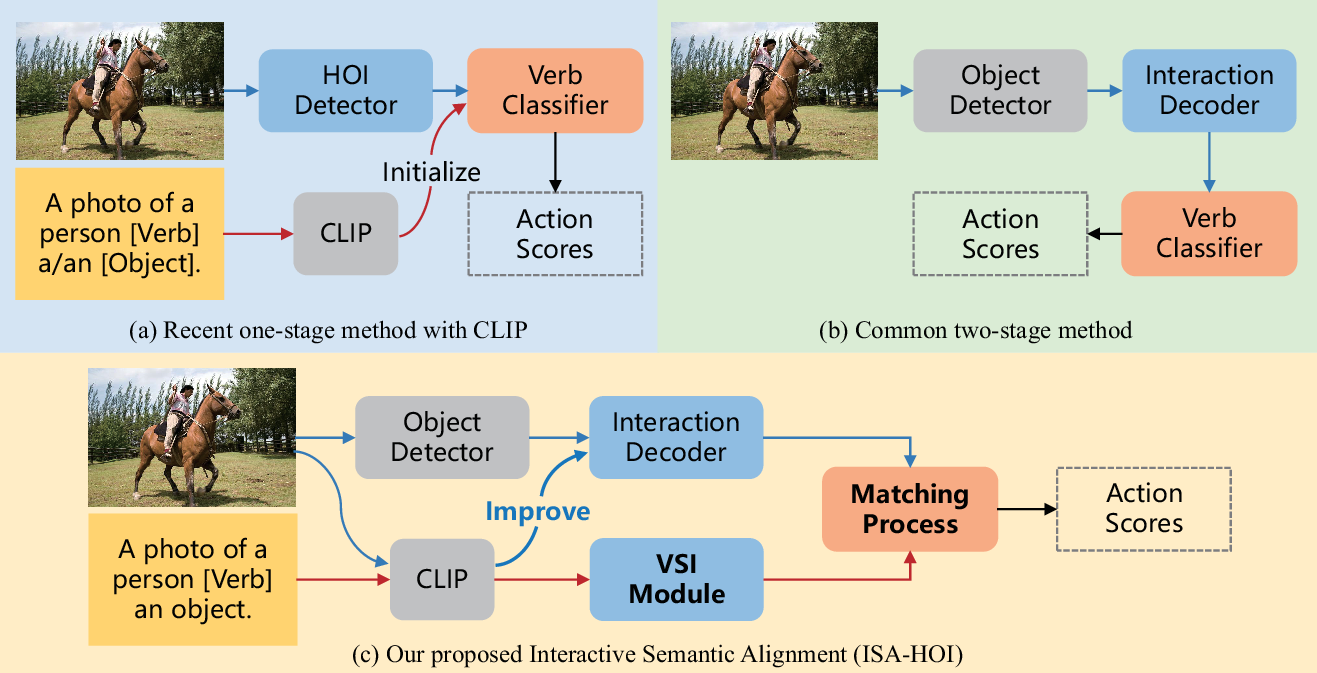} 
	 \caption{\textbf{Comparison of the Architecture with existing works.} (a) Recent one-stage method with CLIP directly initializes classifier with labels' text embeddings. (b) Common two-stage method relies solely on object features for interaction recognition. (c) Our proposed ISA-HOI improves interaction features and verb category labels' text embeddings for alignment. Differentiation is highlighted in bold.}
      \label{fig1} 
\end{figure}
\begin{figure*}[t] 
      \centering
	 \includegraphics[width=6in]{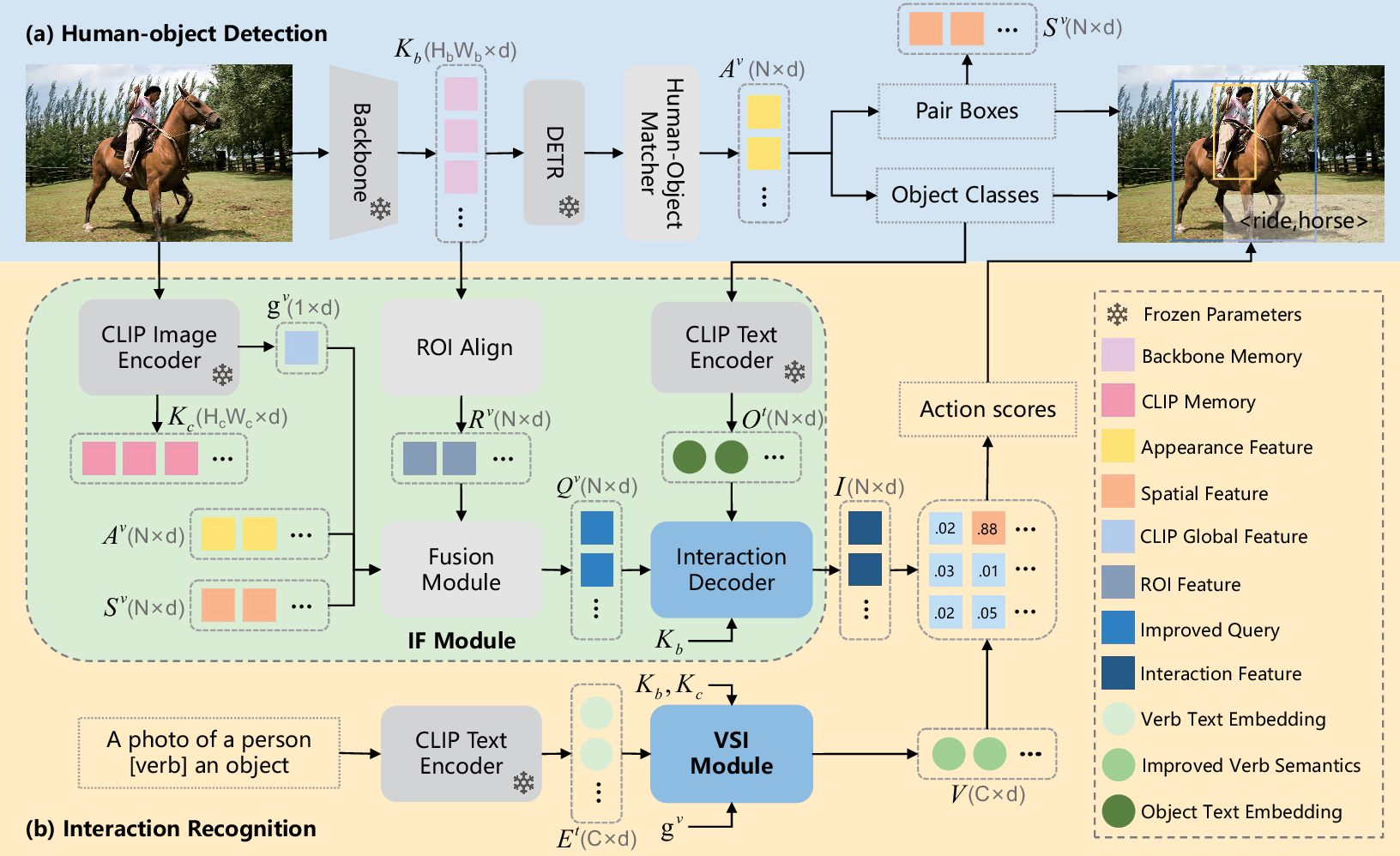} 
	 \caption{\textbf{Overview of our proposed method.} The framework comprises two stages: (a) human-object detection utilizing a pre-trained DETR, and (b) subsequent interaction recognition employing our module IF (Subsec~\ref{IF}) and VSI (Subsec~\ref{VSI}).}
      \label{fig2} 
\end{figure*}

Meanwhile, thanks to the advancement of vision-language models, several studies~\cite{gen,hoiclip,eoid} leverage CLIP~\cite{clip}'s knowledge to delve into zero-shot learning within the HOI tasks. GEN-VLKT~\cite{gen} guided the learned features with image features through knowledge distillation. EoID~\cite{eoid} detected potential action-agnostic interactive human-object pairs by applying a two-stage bipartite matching algorithm and an interactive score module. HOICLIP~\cite{hoiclip} retrieved CLIP knowledge in a query-based manner and harnessed spatial visual features for learning interaction representations. The above methods directly utilize labels’ text embeddings to guide interaction recognition while \textbf{ignoring the heterogeneity} between two different modal features. 

To address the aforementioned limitations, we propose a novel two-stage strategy named ISA-HOI. Unlike previous methods~\cite{gen,pvic} using feed-forward networks (FFN) as interaction classification heads, we consider the recognition of interaction as a matching process between interaction features and verb category labels' text embeddings. Due to the heterogeneity within these two modal features, we design the IF and VSI modules to improve them respectively for alignment. Figure~\ref{fig1} illustrates the straightforward framework of our method and compares it with existing methods.

In the IF module, we utilize the global image features extracted from CLIP to construct interaction queries, which transfers the ability of CLIP to align images and text embeddings to our model while providing contextual information. Specifically, the features extracted from the object detector are concatenated with both the image features obtained from CLIP and the ROI features extracted through ROI-Align~\cite{roi}, thereby containing sufficient visual cues to differentiate interactions. Meanwhile, to prevent spatial features from dominating interaction understanding, we utilize CLIP to obtain predicted object labels' text embeddings, guiding the cross-attention operation within the interaction decoder. 

In the VSI module, we propose a light-weight transformer decoder to improve verb category labels' text embeddings by retrieving knowledge from image features. Through this cross-modal fusion, we further pull verb semantics and interaction features into a unified space.

Our contribution can be summarized as follows:
\begin{itemize}
\item We view interaction recognition as a process that matches interaction features with verb semantics, and efficiently exploit knowledge from pre-trained CLIP model for feature alignment.
\item Our proposed IF module can integrate global and local features while narrowing the distance between interaction features and verb semantics.
\item We further improve the verb category labels' text embeddings via the VSI module, reducing the heterogeneity between them and interaction features.
\end{itemize}
\section{Method}
\subsection{Overview}
Figure~\ref{fig2} illustrates the overall architecture of our proposed two-stage method, which consists of three main components, DETR~\cite{detr} for human-object detection, module IF and VSI for interaction recognition. In the first stage, given an image as input, we employ pre-trained DETR to detect instances and iterate through all valid human-object pairs. The results are represented as $\left\{ \textbf{A}^{v}, \textbf{S}^{v}, (\textbf{B}_h, \textbf{B}_o), \textbf{S}_c, \textbf{O}_{c} \right\}$, where $\textbf{A}^{v}\in \mathbb{R}^{N\times d}$ and $\textbf{S}^{v}\in \mathbb{R}^{N\times d}$ denote appearance features and manually crafted spatial features, $(\textbf{B}_h, \textbf{B}_o)\in \mathbb{R}^{N\times 8}$ denotes bounding boxes, $\textbf{S}_c\in [0, 1]^{N}$ denotes confidence scores, $\textbf{O}_{c}$ denotes a set of N predicted object classes in the format of $<$a photo of a/an \textit{object}$>$, and $N$ denotes the number of pairs.

In the second stage, given a set of C verb categories' text prompts $\textbf{V}_{c}$ in the format of $<$a photo of a person \textit{verb} an object$>$ as additional input, where $C$ is the number of verb categories. Module IF and VSI improve interaction features and verb semantics for feature alignment, respectively. Subsequently, the action scores $\textbf{S}_a\in [0,1]^{N\times C}$ are derived through the cosine similarity between interaction features and verb semantics. Ultimately, $\left\{ (\textbf{B}_h, \textbf{B}_o), \textbf{O}_{c}, \textbf{S}_a\right\}$ collectively form the HOI triplet.

\subsection{Improved Features} \label{IF}
\begin{figure}[t] 
      \centering
	 \includegraphics[width=3in]{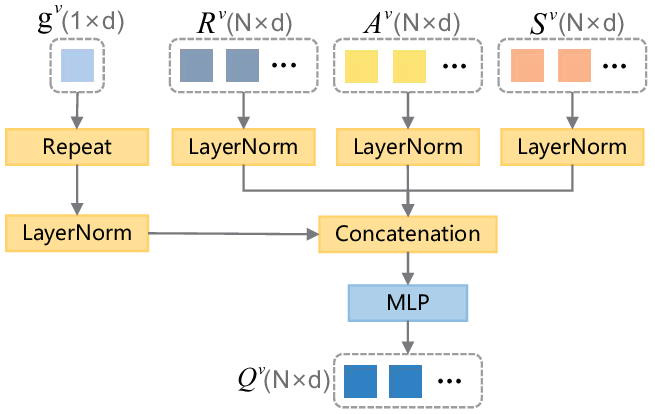} 
	 \caption{The composition of interaction queries.}
      \label{mf} 
\end{figure}
In this subsection, we introduce the architecture of our IF module. we first employ a frozen CLIP image encoder to capture the global token $\textbf{g}^{v}\in \mathbb{R}^{1 \times d}$ of the image, which contains rich contextual information and prior knowledge, significantly enhancing the model's performance as shown in Subsec~\ref{AS}. Simultaneously, we use ROI-Align~\cite{roi} to extract ROI features $\textbf{R}^{v}\in \mathbb{R}^{N \times d}$ for each detected pairs from the feature map. Finally, we utilize a two-layer MLP to fuse all extracted features, thereby obtaining the interaction queries $\textbf{Q}^{v}\in \mathbb{R}^{N\times d}$. Our approach to feature fusion is elaborated in Figure~\ref{mf}. We apply LayerNorm~\cite{ln} before concatenation to prevent numerical overflow.

The interaction decoder utilizes the feature map $\textbf{K}_b$ from the backbone as its keys/values and is guided by the predicted object labels' text embeddings $\textbf{O}^{t}\in \mathbb{R}^{N\times d}$, which 
are derived by encoding the object class $\textbf{O}_c$ using frozen CLIP text encoder. Formally,
\begin{align}
\textbf{O}^{t}&=\textit{TextEnc}(\textbf{O}_c), \\
\textbf{Q}=\textit{MHSA}&(\textbf{Q}^{v}+\textbf{O}^{t}, \textbf{Q}^{v}+\textbf{O}^{t},\textbf{Q}^{v}), \\
\textbf{I}=\textit{MH}&\textit{CA}([\textbf{Q},\textbf{O}^{t}], \textbf{K}_b, \textbf{K}_b), 
\end{align}
where [·,·], $\textit{MHSA}, \textit{MHCA}$ denote concatenation operation, multi-head self-attention and multi-head cross-attention, respectively. $\textbf{I}\in \mathbb{R}^{N\times d}$ represents the final improved interaction features. We exclusively display the essential steps.

\begin{table*}
\begin{center}
\caption{Comparison of Regular HOI detection performance on HICO-DET and V-COCO test sets. } \label{tab:Regular}
\begin{tabular}{llcccccccc}
\toprule
& & &\multicolumn{6}{c}{\textbf{HICO-DET}} & \textbf{V-COCO} \\
& & &\multicolumn{3}{c}{Default Setting} & \multicolumn{3}{c}{Known Objects} & \\
\cmidrule(r){4-6}  \cmidrule(r){7-9}
\textbf{Method}& Backbone & Epochs & Full & Rare & Non-rare & Full & Rare & Non-rare & $AP_{role}^{S2}$ \\
\bottomrule
QPIC~\cite{qpic}\textcolor{black!70}{${\rm _{[2021]}}$} & ResNet-101 & 100 & 29.90 & 23.92 & 31.69 & 32.38 & 26.06 & 34.7 & 61.0 \\
OCN~\cite{ocn}\textcolor{black!70}{${\rm _{[2022]}}$} & ResNet-101 & 80 & 31.43 & 25.80 & 33.11 & - & - & - & 67.1 \\
CDN~\cite{cdn}\textcolor{black!70}{${\rm _{[2021]}}$} & ResNet-101 & 90 & 32.07 & 27.19 & 33.53 & 34.79 & 29.48 & 36.38 & 64.2 \\
DT~\cite{dt}\textcolor{black!70}{${\rm _{[2022]}}$} & ResNet-50 & 80 & 31.75 & 27.45 & 33.03 & 34.50 & 30.13 & 35.81 & \underline{68.5} \\
STIP~\cite{stip}\textcolor{black!70}{${\rm _{[2022]}}$} & ResNet-50 & 30 & 32.22 & 28.15 & 33.43 & 35.29 & 31.43 & 36.45 & \textbf{70.7} \\
UPT~\cite{upt}\textcolor{black!70}{${\rm _{[2022]}}$} & ResNet-101 & 20 & 32.31 & 28.55 & 33.44 & 35.65 & 31.60 & 36.86 & 66.2 \\
RLIP~\cite{rlip}\textcolor{black!70}{${\rm _{[2022]}}$} & ResNet-50 & 90 & 32.84 & 26.85 & 35.10 & - & - & - & 64.2 \\
GEN-VLKT~\cite{gen}\textcolor{black!70}{${\rm _{[2022]}}$} & ResNet-50 & 90 & 33.75 & 29.25 & 35.10 & 36.78 & 32.75 & 37.99 & 64.5 \\
ADA-CM~\cite{ada}\textcolor{black!70}{${\rm _{[2023]}}$} & ResNet-50 & 15 & 33.80 & 31.72 & 34.42 & - & - & - & 61.5\\
HOICLIP~\cite{hoiclip}\textcolor{black!70}{${\rm _{[2023]}}$} & ResNet-50 & 90 & 34.69 & 31.12 & 35.74 & 37.61 & 34.47 & 38.54 & 64.8 \\
PViC~\cite{pvic}\textcolor{black!70}{${\rm _{[2023]}}$} & ResNet-50 & 30 & 34.69 & 32.14 & 35.45 & 38.14 & 35.38 & 38.97 & 65.4 \\
\hline
\rowcolor{gray!25}
ISA-HOI$_s$ & ResNet-50 & 15 & \underline{35.54} & \underline{34.03} & \underline{35.99} & \underline{38.99} & \underline{37.48} & \underline{39.44} & 65.5\\
\rowcolor{gray!25}
ISA-HOI$_l$ & Swin-L & 15 & \textbf{43.66} & \textbf{43.01} & \textbf{43.86} & \textbf{46.52} & \textbf{45.52} & \textbf{46.82} & 67.6\\
\hline
\end{tabular}
\end{center}
\end{table*}

\begin{table}[t]
\begin{center}
\caption{Ablations of different modules of our method on HICO-DET test set under default setting. All variants utilize DETR with ResNet-50 backbone. } \label{tab:ablation}
\begin{tabular}{ccccccc}
\hline
& \multicolumn{3}{c}{IF} & \multicolumn{3}{c}{HICO-DET} \\
\cmidrule(r){2-4}  \cmidrule(r){5-7}
VSI & $g^v$ & $R^v$ & $O^t$ & Full & Rare & Non-Rare \\
\hline
&  &  &  & 32.91 & 27.93 & 34.50 \\
\usym{1F5F8} &  &  &  & 33.65 & 28.80 & 35.10 \\
& \usym{1F5F8} & \usym{1F5F8} & \usym{1F5F8} & 34.76 & 31.55 & 35.72 \\
\usym{1F5F8} & \usym{1F5F8} &  &  & 34.89 & 32.69 & 35.55 \\
\usym{1F5F8} & \usym{1F5F8} & \usym{1F5F8} &  & 35.21 & 32.68 & 35.97 \\
\rowcolor{gray!25}
\usym{1F5F8} & \usym{1F5F8} & \usym{1F5F8} & \usym{1F5F8} & \textbf{35.54} & \textbf{34.03} & \textbf{35.99} \\
\hline
\end{tabular}
\end{center}
\end{table}

\subsection{Verb Semantic Improvement} \label{VSI}
In this subsection, we present the design strategies of queries and keys/values within our VSI module respectively.

\noindent \textbf{Design of Queries.} We initially encode $\textbf{V}_c$ into $ \textbf{E}^{t}=\left\{e_1, e_2, ..., e_C\right\} \in \mathbb{R}^{C\times d} $ and $ \textbf{W} \in \mathbb{R}^{C\times d} $as two learnable embeddings using frozen CLIP text encoder. Subsequently, we utilize the global image features $\textbf{g}^{v}$ to preliminarily improve $\textbf{E}^{t}$. For the i-th category, we obtain its improved embeddings $d_{i}\in \mathbb{R}^{2d}$ as
\begin{eqnarray}
d_i = [e_i\odot \textbf{g}^{v}, e_i], \quad i\in 1, ..., C
\end{eqnarray}
where $\odot$ is the Hadamard product. This operation can integrate the matching capability acquired from the initial CLIP training into the decoder~\cite{zegclip}. Ultimately, we utilize $\textbf{D}^{t}=\left\{d_1, d_2, ..., d_C\right\} \in \mathbb{R}^{C\times 2d}$ as the queries.

\noindent \textbf{Design of Keys/Values.} There are two options for keys and values, which are the feature map $\textbf{K}_b$ from the backbone and the patch embeddings $\textbf{K}_c$ from the CLIP image encoder. We simultaneously mine the interaction information from $\textbf{K}_b$ and $\textbf{K}_c$ through cross-attention operation. Ultimately, we obtain the improved verb semantics $ \textbf{V} \in \mathbb{R}^{C\times d} $ as follows,
\begin{align}
\textbf{D}=&~\textit{M}\textit{HSA}( \textbf{D}^{t}, \textbf{D}^{t}, \textbf{D}^{t}), \\
\textbf{U}=\textit{MHCA}(\textbf{D}&, \textbf{K}_b, \textbf{K}_b)+\textit{MHCA}(\textbf{D}, \textbf{K}_c, \textbf{K}_c), \\
&\textbf{V}=\textbf{W}+\mu \textbf{U},
\end{align}
where $\mu  \in  R$ is a weighting parameter. Note that, we apply learnable linear projection layers on  \textbf{D}$^{t}$ to adjust it to the unified dimension.
\subsection{Training and Inference}
During training, we calculate the cosine similarity between interaction features $\textbf{I}$ and verb semantics $\textbf{V}$ to obtain the final verb prediction scores $\textbf{S}_v\in[0,1]^{N\times C}$ as follows, 
\begin{align}
\textbf{S}_v&=\sigma(\textit{Norm}(\textbf{I})\cdot \textit{Norm}(\textbf{V}^T)),
\end{align}
where $\sigma$ is the sigmoid function and $\textit{Norm}$ represents the normalization operation. Due to the inherent nature of proposal generation, the negative examples vastly outnumber the positive samples. To address this imbalance, we calculate binary focal loss~\cite{focal} between $\textbf{S}_v$ and the binary labels $\textbf{L}_v \in \left\{0,1\right\}^{N \times C}$ for training. Note that, invalid actions for each object are masked out. Formally,
\begin{eqnarray}
\mathcal{L} =\textit{FocalLoss}(\textbf{S}_{v},\textbf{L}_{v}).
\end{eqnarray}

During the inference phase, we merge the confidence scores  $\textbf{S}_c$ of human-object detection and verb prediction scores $\textbf{S}_v$ by utilizing the geometric mean employing the hyperparameter $\lambda\in[0,1]$, as follows,
\begin{eqnarray}
\textbf{S}_a = \textbf{S}_{c}^{1-\lambda}\textbf{S}_{v}^{\lambda}.
\end{eqnarray}
\section{Experiments}
\subsection{Experimental Setting} \label{ES}
\textbf{Datasets.} The primary dataset utilized for experiments is the widely used HICO-DET~\cite{hicodet}, comprising 37,633 training images and 9,546 test images. The annotations encompass 600 categories of HOI triplets, derived from 80 object categories and 117 action categories. Among the 600 HOI categories, there are 138 categories with less than 10 training instances, defined as Rare, and the remaining 462 categories are defined as Non-Rare. we also provide results on V-COCO~\cite{vcoco}, a much smaller dataset comprising 2,533 training images, 2,867 validation images, and 4,946 test images. 

\noindent\textbf{Evaluation Metrics.} Following previous work~\cite{qpic}, we use mean Average Precision (mAP) as the evaluation metric. We define a HOI triplet prediction as a true positive example if the predicted human and object bounding boxes have IoUs larger than 0.5 with the corresponding ground-truth bounding boxes, and the predicted action class is correct. 

\noindent \textbf{Implementation Details.} We employ the DETR~\cite{detr} with ResNet-50~\cite{resnet} backbone and more advanced $\mathcal{H}$-DETR~\cite{hdetr} with Swin-L~\cite{swin} backbone as detectors, named ISA-HOI$_s$ and ISA-HOI$_{l}$ respectively. For CLIP text and image encoder, we employ ViT-B/16 as the backbone and freeze the weights. The interaction decoder and VSI module consist of 2 layers respectively. Within each layer, the embeddings dimension is set to 512, the multi-head attention employs 8 heads, and the hidden dimension of the feed-forward network is configured at 2048. For the focal loss, we set $\alpha$ = 0.5 and $\gamma$ = 0.1. Additionally, the hyper-parameter $\lambda$ is set to 0.26 and the weighting parameter $\mu$ is set to 0.5 and 0.25 respectively under the regular and zero-shot settings.
We train the model for 15 epochs with an initial learning rate of $10^{-4}$ decreased by 5 times at the 10th epoch on 2 Geforce RTX 3090 GPUs.

\noindent \textbf{Data Construction for Zero-Shot Learning.} Following previous work~\cite{gen}, we conduct zero-shot experiments in four manners: Non-rare First Unseen Combination (NF-UC),  Rare First Unseen Combination (RF-UC), Unseen Object (UO) and Unseen Verb (UV). Specifically, within UC setting, training data contains all categories of object and verb but misses 120 HOIs. The NF-UC setting selects unseen categories from head HOIs, while the RF-UC setting indicates the tail HOIs are selected as unseen classes. Within the UO setting, we construct unseen HOIs consisting of 12 unseen objects from the total 80 objects, creating a set of 100 unseen and 500 seen HOIs. Within the UV setting, we select 20 verbs from the total of 117 verbs to constitute 84 unseen and 516 seen HOIs.
\begin{table}[t]
\begin{center}
\caption{Component analysis of VSI module on HICO-DET test set under default setting.} \label{tab:VSI}
\begin{tabular}{cccc}
\hline
Strategy & Full & Rare & Non-Rare \\
\hline
\rowcolor{gray!25}
ISA-HOI$_{s}$ & \textbf{35.54} & \textbf{34.03} & 35.99 \\
w/o $g^v$    & 35.24 & 32.65 & 36.02 \\
w/o $K_b$    & 35.09 & 32.04 & 36.00 \\
w/o $K_c$    & 35.23 & 32.51 & \textbf{36.04} \\
\hline
\end{tabular}
\end{center}
\end{table}
\begin{table}[t!]
\begin{center}
\caption{Performance comparison with different layer number of the VSI Module. All variants are based on ISA-HOI$_s$.} \label{tab:num}
\begin{tabular}{cccc}
\hline
Layer number & Full & Rare & Non-Rare \\
\hline
0 & 34.61 & 32.02 & 35.39 \\
1 & 35.11 & 32.85 & 35.78 \\
\rowcolor{gray!25}
2 & \textbf{35.54} & \textbf{34.03} & \textbf{35.99} \\
3    & 35.38 & 33.36 & 35.98 \\
\hline
\end{tabular}
\end{center}
\end{table}
\subsection{Effectiveness for Regular HOI Detection} \label{RHOI}
As shown in Table \ref{tab:Regular}, we show the comparison of HOI detection performance computed by official evaluation code on HICO-DET and V-COCO test sets. We showcase our model's performance using two different detectors to illustrate its scalability. For HICO-DET dataset, ISA-HOI$_s$ surpasses the previous state-of-the-art two-stage method PViC~\cite{pvic} by 0.85 mAP while using only half the training epochs. Moreover, compared to GEN-VLKT~\cite{gen} and HOICLIP~\cite{hoiclip}, which also extract prior knowledge from CLIP~\cite{clip}, our method outperforms them by 1.79 and 0.85 mAP respectively. Noteworthy, ISA-HOI$_l$ experiences a significant performance boost when utilizing a stronger detector. For V-COCO dataset, our method also demonstrates competitive performance, although it primarily emphasizes actions, which occasionally results in annotations lacking object annotations.

\begin{table}[t]
\begin{center}
\caption{Comparison of Zero-shot HOI detection performance on HICO-DET test set.} \label{tab:zero-shot}
\begin{tabular}{lcccc}
\hline
Method & Type & Unseen & Seen & Full \\
\hline
FCL~\cite{fcl}\textcolor{black!70}{${\rm _{[2021]}}$} & NF-UC & 18.66 & 19.55 & 19.37 \\
GEN-VLKT~\cite{gen}\textcolor{black!70}{${\rm _{[2022]}}$} & NF-UC & 25.05 & 23.38 & 23.71 \\
EoID~\cite{eoid}\textcolor{black!70}{${\rm _{[2023]}}$} & NF-UC & 26.77 & 26.66 & 26.69 \\
HOICLIP~\cite{hoiclip}\textcolor{black!70}{${\rm _{[2023]}}$} & NF-UC & 26.39 & 28.10 & 27.75 \\
\rowcolor{gray!25}
ISA-HOI$_s$ & NF-UC & \underline{32.30} & \underline{28.92} & \underline{29.60} \\
\rowcolor{gray!25}
ISA-HOI$_l$ & NF-UC & \textbf{38.77} & \textbf{37.36} & \textbf{37.64} \\
\hline
FCL~\cite{fcl}\textcolor{black!70}{${\rm _{[2021]}}$} & RF-UC & 13.16 & 24.23 & 22.01 \\
GEN-VLKT~\cite{gen}\textcolor{black!70}{${\rm _{[2022]}}$} & RF-UC & 21.36 & 32.91 & 30.56 \\
EoID~\cite{eoid}\textcolor{black!70}{${\rm _{[2023]}}$} & RF-UC & 22.04 & 31.39 & 29.56 \\
HOICLIP~\cite{hoiclip}\textcolor{black!70}{${\rm _{[2023]}}$} & RF-UC & \underline{25.53} & \underline{34.85} & \underline{32.99} \\

\rowcolor{gray!25}
ISA-HOI$_s$ & RF-UC & 24.96 & 34.72 & 32.77 \\
\rowcolor{gray!25}
ISA-HOI$_l$ & RF-UC & \textbf{26.68} & \textbf{41.34} & \textbf{38.41} \\
\hline
FCL~\cite{fcl}\textcolor{black!70}{${\rm _{[2021]}}$} & UO & 15.54 & 20.74 & 19.87 \\
GEN-VLKT~\cite{gen}\textcolor{black!70}{${\rm _{[2022]}}$} & UO & 10.51 & 28.92 & 25.63  \\
HOICLIP~\cite{hoiclip}\textcolor{black!70}{${\rm _{[2023]}}$} & UO & 16.20 & 30.99 & 28.53 \\
\rowcolor{gray!25}
ISA-HOI$_s$ & UO & \underline{30.10} & \underline{31.40} & \underline{31.19} \\
\rowcolor{gray!25}
ISA-HOI$_l$ & UO & \textbf{34.92} & \textbf{39.29} & \textbf{38.56} \\
\hline
GEN-VLKT~\cite{gen}\textcolor{black!70}{${\rm _{[2022]}}$} & UV & 20.96 & 30.23 & 28.74  \\
EoID~\cite{eoid}\textcolor{black!70}{${\rm _{[2023]}}$} & UV & 22.71 & 30.73 & 29.61 \\
HOICLIP~\cite{hoiclip}\textcolor{black!70}{${\rm _{[2023]}}$} & UV & 24.30 & 31.14 & 30.42 \\
\rowcolor{gray!25}
ISA-HOI$_s$ & UV & \underline{25.66} & \underline{32.07} & \underline{31.17} \\
\rowcolor{gray!25}
ISA-HOI$_l$ & UV & \textbf{28.05} & \textbf{40.86} & \textbf{39.07} \\
\hline
\end{tabular}
\end{center}
\end{table}

\subsection{Ablation Study} \label{AS}
As shown in Table~\ref{tab:ablation}, we conduct a series of experiments to analyze the effectiveness of our proposed modules. Initially, we introduce a baseline model by emulating the design of PViC~\cite{pvic} but omit the positional embeddings for guidance. We first introduce our VSI module, which can enable labels' text embeddings to fully mine knowledge from image features for alignment, which achieves a 0.74 mAP improvement in full categories. Meanwhile, we separately add the IF module to improve interaction features, which brings a 1.85 mAP improvement. Next, we gradually incorporate different features to form the IF module, image features from CLIP, ROI features and predicted object labels' text embeddings improve the performance by 1.24 mAP, 0.32 mAP and 0.33 mAP in full categories, respectively. We have observed that the introduction of CLIP features greatly improves the performance especially for rare categories, as it can utilize CLIP's prior knowledge to narrow the distance between interaction features and verb semantics while providing rich global information. 

Additionally, we investigate the improvement effect of different image features on verb semantics, as shown in Figure~\ref{tab:VSI}. The inclusion of image features significantly boosts the model's recognition capability in rare categories. To explore the effect of layer number $L$ in VSI module, we show the performances on HICO-DET test set under default setting by varying $L$ from 0 to 3. As shown in Table~\ref{tab:num}, the best performances are achieved when $L$ is set to 2.

\subsection{Effectiveness for Zero-Shot HOI Detection} \label{ZHOI}
As shown in Table~\ref{tab:zero-shot}, we conduct experiments under four zero-shot settings following previous work~\cite{gen}: Non-Rare First Unseen Composition (NF-UC), Rare First Unseen Composition (RF-UC), Unseen Object (UO) and Unseen Verb (UV) settings. 
Under the NF-UC setting, ISA-HOI$_s$ promotes mAP from 26.77 to 32.30 compared to EoID~\cite{eoid} in unseen categories, which fully demonstrates the ability to predict unseen interactions. For the UO setting, which mirrors the capacity to explore interactions with unseen objects, our model promotes performance by a significant margin of 14.9 mAP compared to HOICLIP. For RF-UC and UV settings, our model also performs excellently, with performance comparable to the state-of-the-art method. Likewise, we conduct experiments utilizing ISA-HOI$_l$ under zero-shot settings, revealing significant improvements in the results. Note that, under zero-shot settings, we modify the number of predicted categories of model to 600 HOIs instead of 117 actions for fair comparison.

\subsection{Visualization} \label{visual}
\begin{figure}[t]
  \centering
  \subfigure[Interactions involving the same individual but different objects.]{
    \begin{minipage}[t]{1in}
        \centering
        \includegraphics[width=1in]{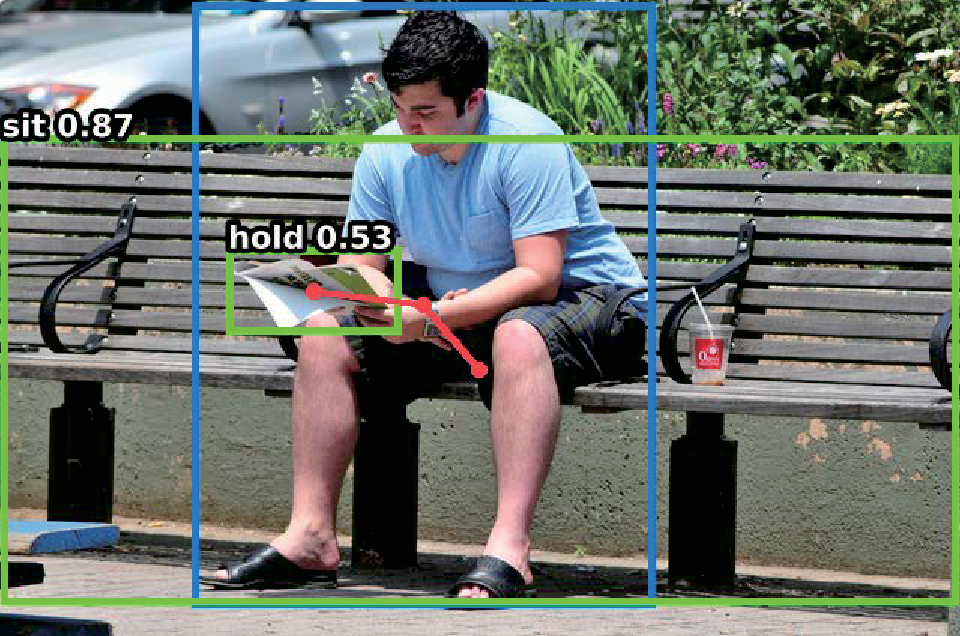}
    \end{minipage}
    \begin{minipage}[t]{1in}
        \centering
        \includegraphics[width=1in]{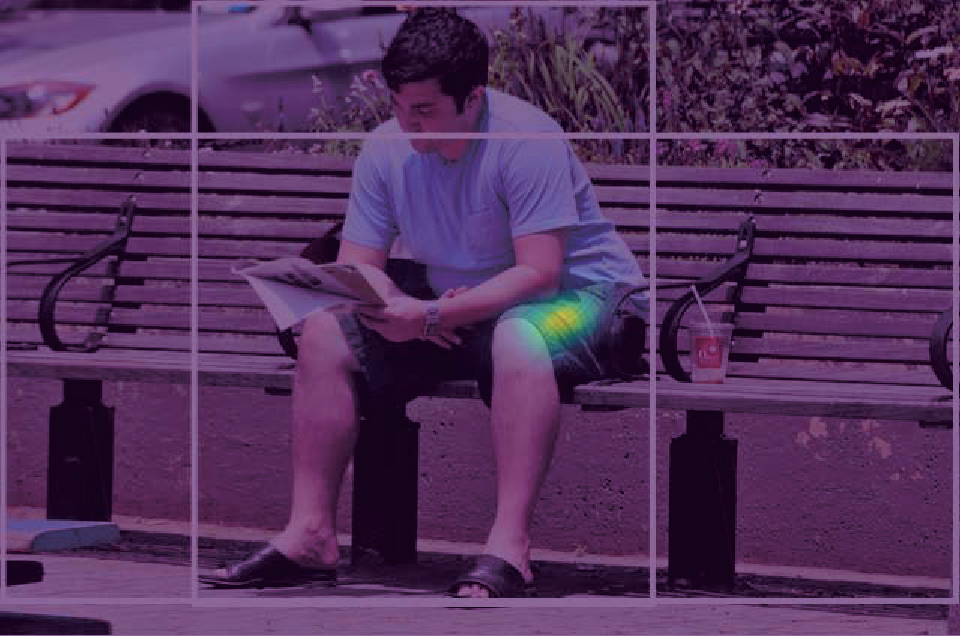}
    \end{minipage}
    \begin{minipage}[t]{1in}
        \centering
        \includegraphics[width=1in]{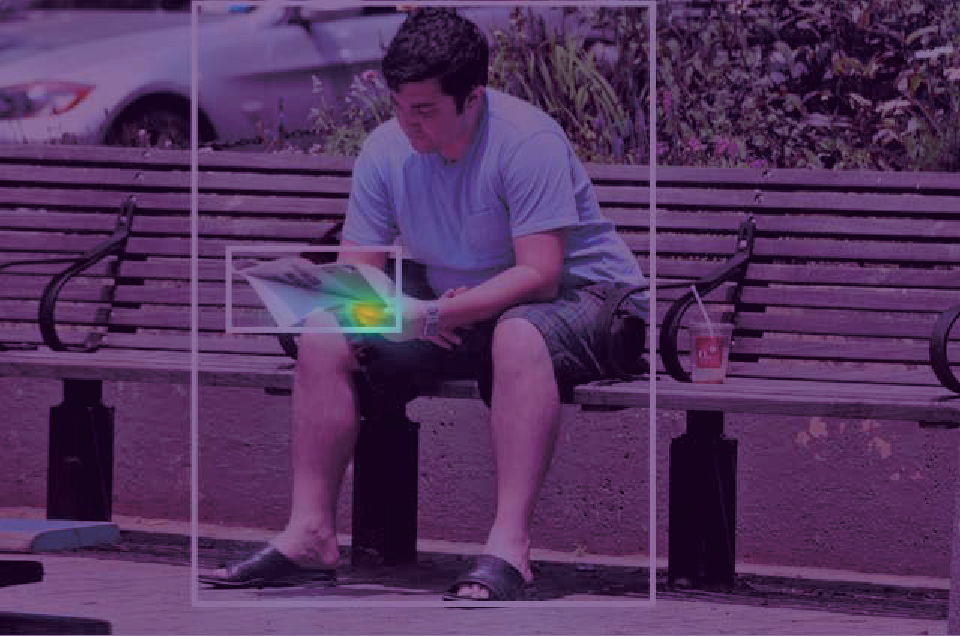}
    \end{minipage}
  }
    \subfigure[Interactions involving the same object but different individuals.]{
    \begin{minipage}[t]{1in}
        \centering
        \includegraphics[width=1in]{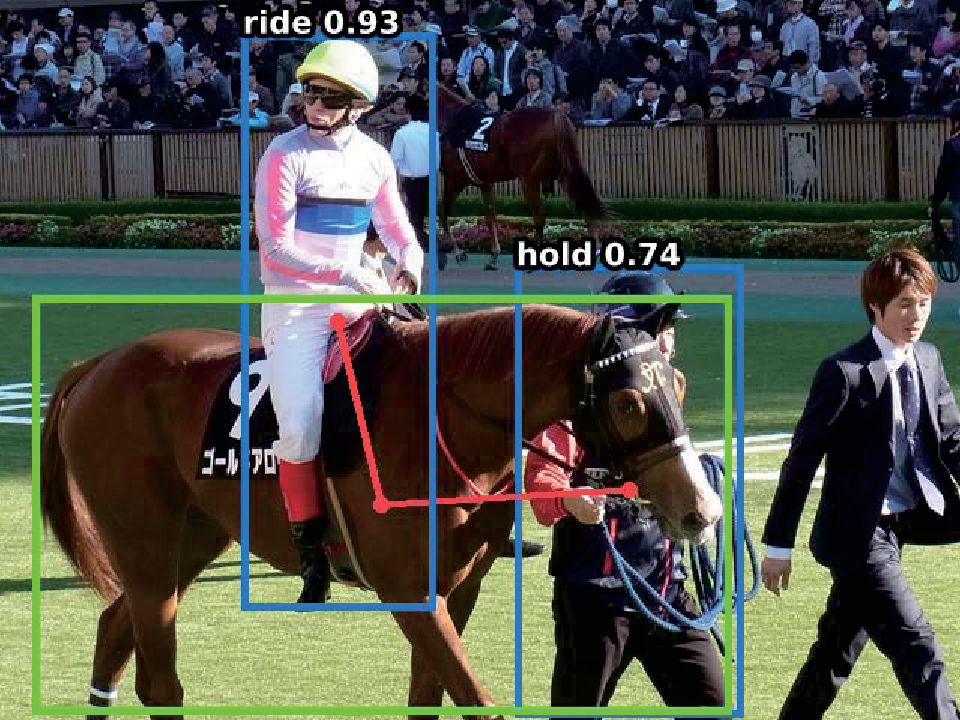}
    \end{minipage}
    \begin{minipage}[t]{1in}
        \centering
        \includegraphics[width=1in]{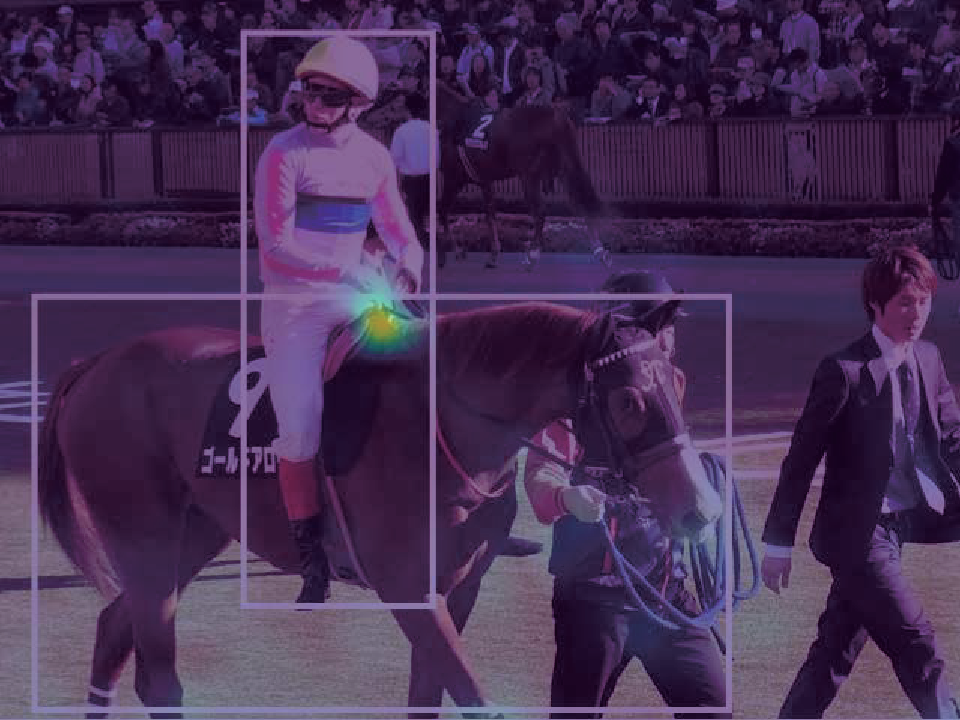}
    \end{minipage}
    \begin{minipage}[t]{1in}
        \centering
        \includegraphics[width=1in]{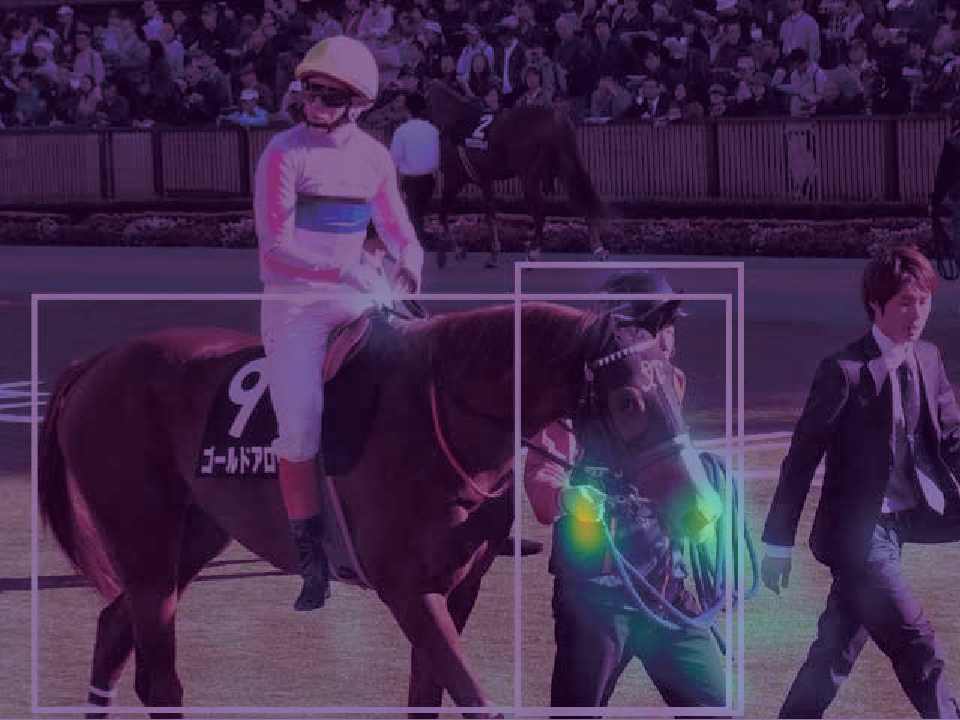}
    \end{minipage}
  }
  
    \subfigure[Interactions involving different individuals and objects.~~~~~~~~~~~~]{
    \begin{minipage}[t]{1in}
        \centering
        \includegraphics[width=1in]{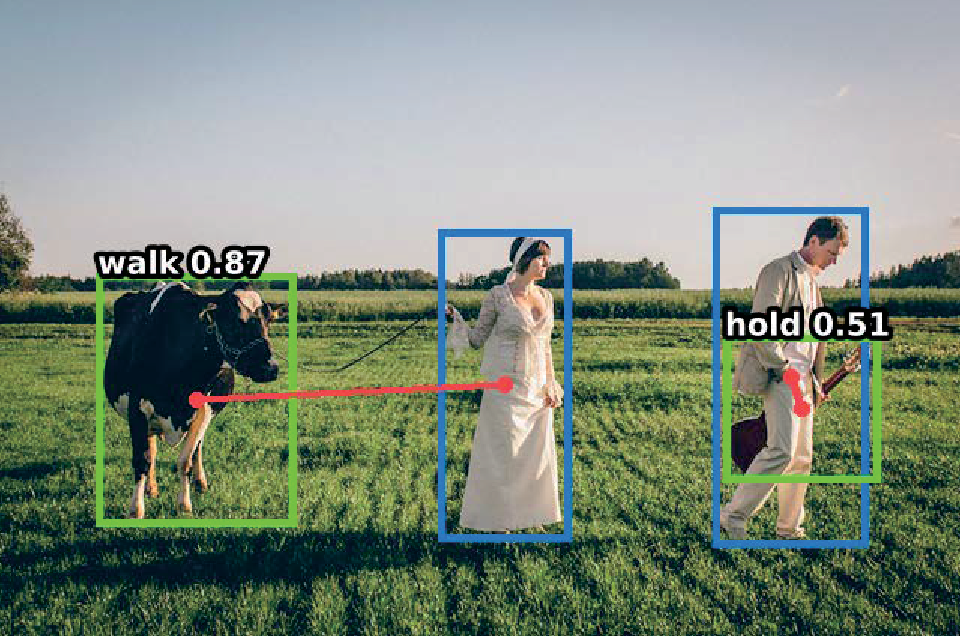}
    \end{minipage}
    \begin{minipage}[t]{1in}
        \centering
        \includegraphics[width=1in]{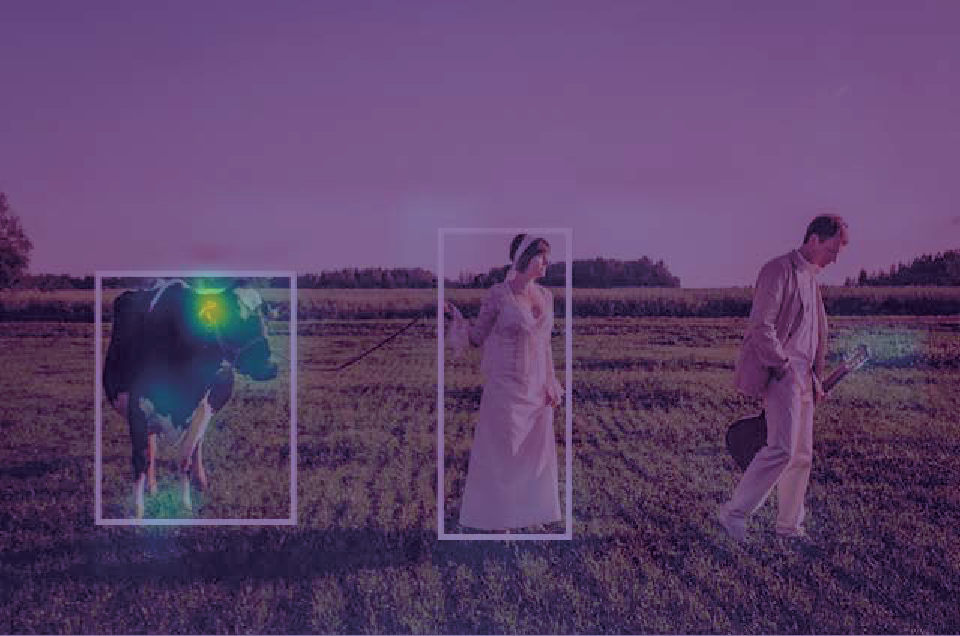}
    \end{minipage}
    \begin{minipage}[t]{1in}
        \centering
        \includegraphics[width=1in]{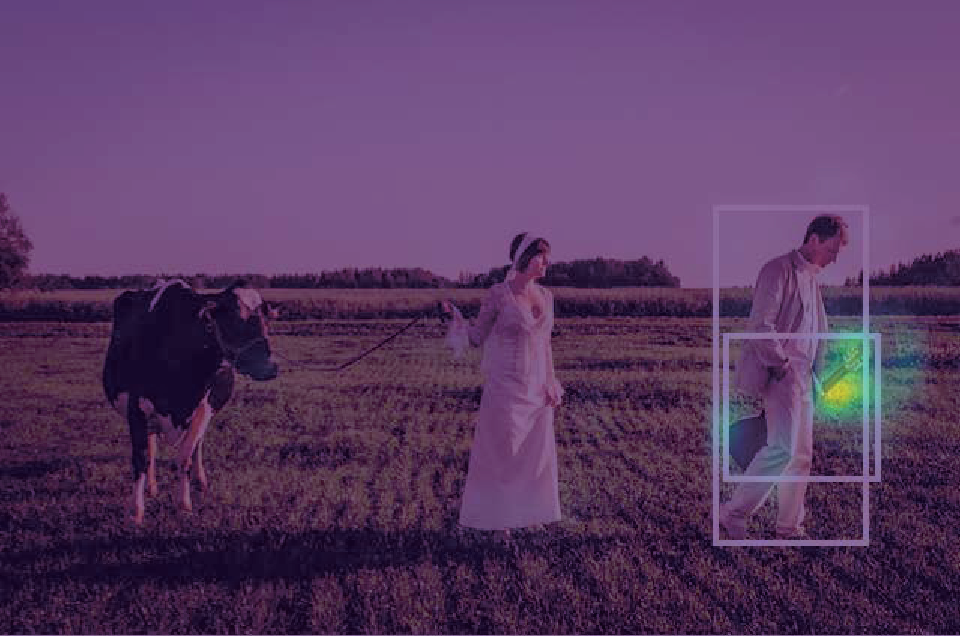}
    \end{minipage}
  }
  \caption{\textbf{Visualization of predictions.} The first column displays the predicted results of ISA-HOI$_{s}$, while the second and third columns showcase attention maps of various interactions. All images are sampled from the HICO-DET test set.}
  \label{vis} 
\end{figure}

\begin{figure}[t] 
      \centering
	 \includegraphics[width=2in]{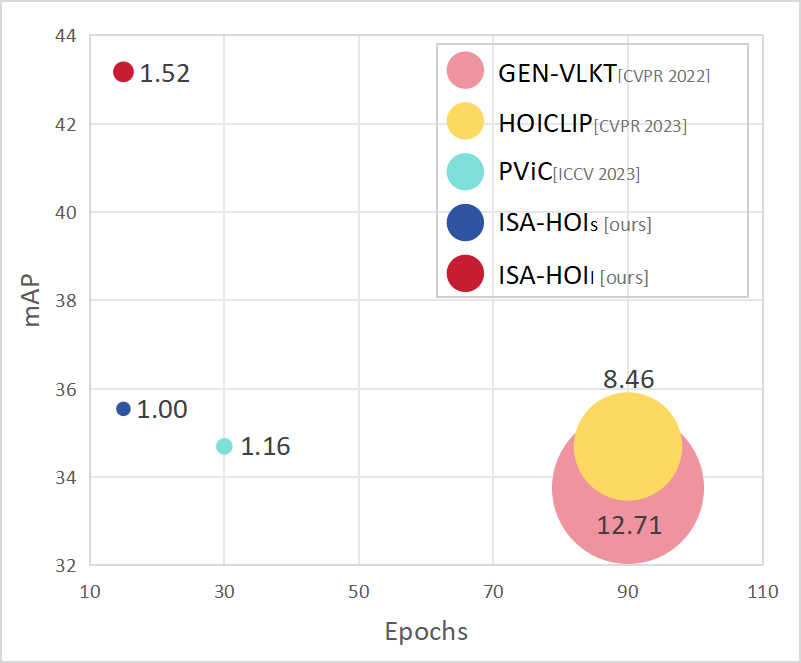} 
	 \caption{\textbf{Visualization of efficiency comparison.} The size of the points directly proportional to the relative training gpu time annotated alongside them.}
      \label{eff} 
\end{figure}

As shown in Figure~\ref{vis}, we visualize the prediction results and attention maps of our model ISA-HOI$_s$. Across different interactions within the same image, ISA-HOI$_s$ effectively concentrates on specific regions, yielding accurate prediction results. Furthermore, we conduct a comparison of the training efficiency among various methods, as depicted in Figure~\ref{eff}. Overall, two-stage methods exhibit superiority over one-stage methods owing to the ability to utilize the pre-trained object detector. Specifically, our method demonstrates the most superior efficiency performance due to the improved queries that elevate the efficiency of the interaction decoder. 

\section{Conclusion}
We present a novel strategy named ISA-HOI, exploiting prior knowledge from CLIP and aligning interactive semantics between interaction features in images and textual features of verb categories. Extensive experiments demonstrate the superiority of our method. Moving forward, our target is to investigate the relationship between humans, objects and actions, aiming to improve the model's capability to recognize unseen HOI categories.
\section*{Acknowledgment}
This research was partly supported by grants of National Natural Science Foundation of China (NSFC, Grant No. 62171281), Science and Technology Commission of Shanghai Municipality (STCSM, Grant Nos. 20DZ1200203, 2021SHZDZX0102, 22DZ2229005).
\bibliographystyle{IEEEbib}
\bibliography{icme24_camera_ready}

\begin{thebibliography}{10}

\bibitem{scg}
Frederic~Z Zhang, Dylan Campbell, and Stephen Gould,
\newblock ``Spatially conditioned graphs for detecting human-object interactions,''
\newblock in {\em Proceedings of the IEEE/CVF International Conference on Computer Vision}, 2021, pp. 13319--13327.

\bibitem{stip}
Yong Zhang, Yingwei Pan, Ting Yao, Rui Huang, Tao Mei, and Chang-Wen Chen,
\newblock ``Exploring structure-aware transformer over interaction proposals for human-object interaction detection,''
\newblock in {\em Proceedings of the IEEE/CVF Conference on Computer Vision and Pattern Recognition}, 2022, pp. 19548--19557.

\bibitem{upt}
Frederic~Z Zhang, Dylan Campbell, and Stephen Gould,
\newblock ``Efficient two-stage detection of human-object interactions with a novel unary-pairwise transformer,''
\newblock in {\em Proceedings of the IEEE/CVF Conference on Computer Vision and Pattern Recognition}, 2022, pp. 20104--20112.

\bibitem{pvic}
Frederic~Z Zhang, Yuhui Yuan, Dylan Campbell, Zhuoyao Zhong, and Stephen Gould,
\newblock ``Exploring predicate visual context in detecting of human-object interactions,''
\newblock in {\em Proceedings of the IEEE/CVF International Conference on Computer Vision}, 2023, pp. 10411--10421.

\bibitem{qpic}
Masato Tamura, Hiroki Ohashi, and Tomoaki Yoshinaga,
\newblock ``Qpic: Query-based pairwise human-object interaction detection with image-wide contextual information,''
\newblock in {\em Proceedings of the IEEE/CVF Conference on Computer Vision and Pattern Recognition}, 2021, pp. 10410--10419.

\bibitem{cdn}
Aixi Zhang, Yue Liao, Si~Liu, Miao Lu, Yongliang Wang, Chen Gao, and Xiaobo Li,
\newblock ``Mining the benefits of two-stage and one-stage hoi detection,''
\newblock {\em Advances in Neural Information Processing Systems}, vol. 34, pp. 17209--17220, 2021.

\bibitem{gen}
Yue Liao, Aixi Zhang, Miao Lu, Yongliang Wang, Xiaobo Li, and Si~Liu,
\newblock ``Gen-vlkt: Simplify association and enhance interaction understanding for hoi detection,''
\newblock in {\em Proceedings of the IEEE/CVF Conference on Computer Vision and Pattern Recognition}, 2022, pp. 20123--20132.

\bibitem{hoiclip}
Shan Ning, Longtian Qiu, Yongfei Liu, and Xuming He,
\newblock ``Hoiclip: Efficient knowledge transfer for hoi detection with vision-language models,''
\newblock in {\em Proceedings of the IEEE/CVF Conference on Computer Vision and Pattern Recognition}, 2023, pp. 23507--23517.

\bibitem{attention}
Ashish Vaswani, Noam Shazeer, Niki Parmar, Jakob Uszkoreit, Llion Jones, Aidan~N Gomez, {\L}ukasz Kaiser, and Illia Polosukhin,
\newblock ``Attention is all you need,''
\newblock {\em Advances in Neural Information Processing Systems}, vol. 30, 2017.

\bibitem{eoid}
Mingrui Wu, Jiaxin Gu, Yunhang Shen, Mingbao Lin, Chao Chen, and Xiaoshuai Sun,
\newblock ``End-to-end zero-shot hoi detection via vision and language knowledge distillation,''
\newblock in {\em Proceedings of the AAAI Conference on artificial intelligence}, 2023, vol.~37, pp. 2839--2846.

\bibitem{clip}
Alec Radford, Jong~Wook Kim, Chris Hallacy, Aditya Ramesh, Gabriel Goh, Sandhini Agarwal, Girish Sastry, Amanda Askell, Pamela Mishkin, Jack Clark, et~al.,
\newblock ``Learning transferable visual models from natural language supervision,''
\newblock in {\em International conference on machine learning}. PMLR, 2021, pp. 8748--8763.

\bibitem{roi}
Kaiming He, Georgia Gkioxari, Piotr Doll{\'a}r, and Ross Girshick,
\newblock ``Mask r-cnn,''
\newblock in {\em Proceedings of the IEEE/CVF International Conference on Computer Vision}, 2017, pp. 2961--2969.

\bibitem{detr}
Nicolas Carion, Francisco Massa, Gabriel Synnaeve, Nicolas Usunier, Alexander Kirillov, and Sergey Zagoruyko,
\newblock ``End-to-end object detection with transformers,''
\newblock in {\em European conference on computer vision}. Springer, 2020, pp. 213--229.

\bibitem{ln}
Jimmy~Lei Ba, Jamie~Ryan Kiros, and Geoffrey~E Hinton,
\newblock ``Layer normalization,''
\newblock {\em arXiv preprint arXiv:1607.06450}, 2016.

\bibitem{ocn}
Hangjie Yuan, Mang Wang, Dong Ni, and Liangpeng Xu,
\newblock ``Detecting human-object interactions with object-guided cross-modal calibrated semantics,''
\newblock in {\em Proceedings of the AAAI Conference on artificial intelligence}, 2022, vol.~36, pp. 3206--3214.

\bibitem{dt}
Desen Zhou, Zhichao Liu, Jian Wang, Leshan Wang, Tao Hu, Errui Ding, and Jingdong Wang,
\newblock ``Human-object interaction detection via disentangled transformer,''
\newblock in {\em Proceedings of the IEEE/CVF Conference on Computer Vision and Pattern Recognition}, 2022, pp. 19568--19577.

\bibitem{rlip}
Hangjie Yuan, Jianwen Jiang, Samuel Albanie, Tao Feng, Ziyuan Huang, Dong Ni, and Mingqian Tang,
\newblock ``Rlip: Relational language-image pre-training for human-object interaction detection,''
\newblock {\em Advances in Neural Information Processing Systems}, vol. 35, pp. 37416--37431, 2022.

\bibitem{ada}
Ting Lei, Fabian Caba, Qingchao Chen, Hailin Jin, Yuxin Peng, and Yang Liu,
\newblock ``Efficient adaptive human-object interaction detection with concept-guided memory,''
\newblock in {\em Proceedings of the IEEE/CVF International Conference on Computer Vision}, 2023, pp. 6480--6490.

\bibitem{zegclip}
Ziqin Zhou, Yinjie Lei, Bowen Zhang, Lingqiao Liu, and Yifan Liu,
\newblock ``Zegclip: Towards adapting clip for zero-shot semantic segmentation,''
\newblock in {\em Proceedings of the IEEE/CVF Conference on Computer Vision and Pattern Recognition}, 2023, pp. 11175--11185.

\bibitem{focal}
Tsung-Yi Lin, Priya Goyal, Ross Girshick, Kaiming He, and Piotr Doll{\'a}r,
\newblock ``Focal loss for dense object detection,''
\newblock in {\em Proceedings of the IEEE/CVF International Conference on Computer Vision}, 2017, pp. 2980--2988.

\bibitem{hicodet}
Yu-Wei Chao, Yunfan Liu, Xieyang Liu, Huayi Zeng, and Jia Deng,
\newblock ``Learning to detect human-object interactions,''
\newblock in {\em 2018 ieee winter conference on applications of computer vision (wacv)}, 2018, pp. 381--389.

\bibitem{vcoco}
Saurabh Gupta and Jitendra Malik,
\newblock ``Visual semantic role labeling,''
\newblock {\em arXiv preprint arXiv:1505.04474}, 2015.

\bibitem{resnet}
Kaiming He, Xiangyu Zhang, Shaoqing Ren, and Jian Sun,
\newblock ``Deep residual learning for image recognition,''
\newblock in {\em Proceedings of the IEEE/CVF Conference on Computer Vision and Pattern Recognition}, 2016, pp. 770--778.

\bibitem{hdetr}
Ding Jia, Yuhui Yuan, Haodi He, Xiaopei Wu, Haojun Yu, Weihong Lin, Lei Sun, Chao Zhang, and Han Hu,
\newblock ``Detrs with hybrid matching,''
\newblock in {\em Proceedings of the IEEE/CVF Conference on Computer Vision and Pattern Recognition}, 2023, pp. 19702--19712.

\bibitem{swin}
Ze~Liu, Yutong Lin, Yue Cao, Han Hu, Yixuan Wei, Zheng Zhang, Stephen Lin, and Baining Guo,
\newblock ``Swin transformer: Hierarchical vision transformer using shifted windows,''
\newblock in {\em Proceedings of the IEEE/CVF International Conference on Computer Vision}, 2021, pp. 10012--10022.

\bibitem{fcl}
Zhi Hou, Baosheng Yu, Yu~Qiao, Xiaojiang Peng, and Dacheng Tao,
\newblock ``Detecting human-object interaction via fabricated compositional learning,''
\newblock in {\em Proceedings of the IEEE/CVF Conference on Computer Vision and Pattern Recognition}, 2021, pp. 14646--14655.

\end{thebibliography}

\end{document}